\documentclass{article}
\usepackage{ijcai20}

\usepackage{times}
\usepackage{soul}
\usepackage{url}
\usepackage[hidelinks]{hyperref}
\usepackage[utf8]{inputenc}
\usepackage[small]{caption}
\usepackage{graphicx}
\usepackage{amsmath}
\usepackage{amsthm}
\usepackage{booktabs}
\usepackage{algorithm}
\usepackage{algorithmic}
\usepackage{multirow}
\usepackage{bm}
\usepackage{amssymb}
\usepackage{threeparttable}
\usepackage{footnote}
\usepackage{color}
\usepackage{makecell}
\usepackage{float}
\usepackage{placeins}

\title{Split to Be Slim: An Overlooked Redundancy in Vanilla Convolution}

\author{
Qiulin Zhang$^1$
\and
Zhuqing Jiang$^1$\footnote{Corresponding author.}\and
Qishuo Lu$^1$\and
Jia'nan Han$^1$\and\\
Zhengxin Zeng$^1$\and
Shang-Hua Gao$^2$\and
Aidong Men$^1$
\affiliations
$^1$Beijing University of Posts and Telecommunications\\
$^2$Nankai University\\
\emails
\{qiulinzhang, jiangzhuqing, hanjianan, zengzhengxinsice, menad\}@bupt.edu.cn \and \\
shgao@mail.nankai.edu.cn
}

\begin{document}

\maketitle

\begin{abstract}
%
Many effective solutions have been proposed to reduce the redundancy of models for inference acceleration.
Nevertheless, common approaches mostly focus on eliminating less important filters or constructing efficient operations,
while ignoring the pattern redundancy in feature maps.
We reveal that many feature maps within a layer share similar but not identical patterns.
However, it is difficult to identify if features with similar patterns are redundant or contain essential details.
Therefore, instead of directly removing uncertain redundant features,
we propose a \textbf{sp}lit based \textbf{conv}olutional operation, namely SPConv, to tolerate features with similar patterns but require less computation.
Specifically, we split input feature maps into the representative part and the uncertain redundant part,
where intrinsic information is extracted from the representative part through relatively heavy computation while
tiny hidden details in the uncertain redundant part are processed with some light-weight operation.
To recalibrate and fuse these two groups of processed features, we propose a parameters-free feature fusion module.
Moreover, our SPConv is formulated to replace the vanilla convolution in a plug-and-play way.
Without any bells and whistles, experimental results on benchmarks demonstrate SPConv-equipped networks consistently outperform state-of-the-art baselines in both accuracy and inference time on GPU, with FLOPs and parameters dropped sharply.
\end{abstract}

\section{Introduction}
\begin{figure}
  \centering
  \includegraphics[height=6cm]{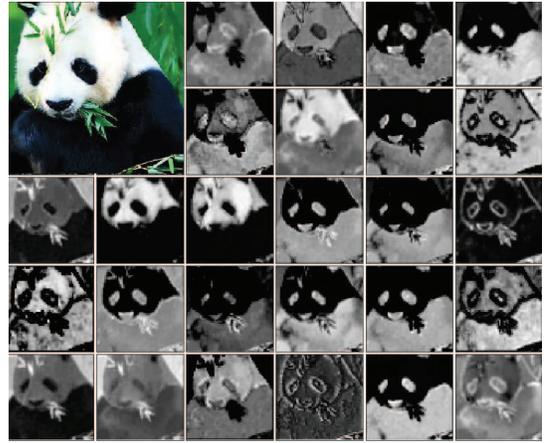}
  \caption{Visualization of input image (top-left) and some input feature maps of the second stage in ResNet-50. Many of them appear much pattern similarity. Therefore, some representative feature maps can be chosen to supplement intrinsic information, while the remaining redundant only need to complement tiny different details.}\label{panda}
\vspace{-0.5cm}
\end{figure}

The recent years have seen the remarkable success of deep neural networks. However, this booming accuracy comes at the cost of increasingly complex models, which contain millions of parameters and billions of FLOPs \cite{DBLP:conf/cvpr/HeZRS16,DBLP:conf/naacl/DevlinCLT19}. Therefore, recent efforts try to reduce parameters and FLOPs while ensuring the model without a significant performance drop.

One main effort is to design an efficient convolutional operation. The popular trend recently tends to make full use of group-wise convolution (GWC) \cite{krizhevsky2012imagenet}, depth-wise convolution (DWC) \cite{vanhoucke2014learning} and point-wise convolution (PWC) \cite{szegedy2015going}. Many  effective and efficient models have been produced by combining these three schemes carefully, such as MobileNet \cite{howard2017mobilenets}, Xception \cite{DBLP:conf/cvpr/Chollet17}, ResNeXt \cite{xie2017aggregated} and ShuffleNet \cite{zhang2018shufflenet}. The idea behind these famous nets illustrates that redundancy in inter-channel connectivity can be reduced reasonably, or to say, the effect of dense vanilla convolutional kernels can be achieved with combinations of some sparse ones.

Different from the pipelines reviewed above, we take a step back and look at inputs. Visualization of the second stage's inputs of ResNet-50 is shown in Figure \ref{panda}. We notice that some feature maps appear much pattern similarity. A natural thought to this phenomenon is that many feature maps are redundant and they can be deleted so that fewer filters are needed for feature extraction during training. However, it is difficult to identify how many channels are economically enough in each layer for different tasks and datasets. The widely used stereotype of 64-128-256-512-1024 channels in each layer/stage, also what we inherit for granted from classic architectures \cite{DBLP:conf/cvpr/HeZRS16}, gives a compromised answer: to be redundant rather than to be in shortage for better performance on most of the datasets. This stereotype implies it is very tough to obtain the exact number of channels. Therefore, instead of struggling with such an obstacle, we incline to tolerate the redundancy among feature maps and treat them with a more elegant method. Back to Figure \ref{panda}, some of the feature maps appear very similar, which indicates the information they carry is also similar. However, in vanilla convolution, similar feature maps are convolved repetitively by different $k\times k \times 1$ kernels. Intuitively, if one model has extracted information from one feature map, it will only need to extract the difference from its similar ones.

Inspired by the idea above, in this paper, we propose a novel SPConv module to reduce the redundancy in vanilla convolution. Specifically, we split all input channels into two parts: one for the representative and the other for the redundant. Intrinsic information can be extracted from the representative part by normal $k\times k$ kernels. Complementarily, the hidden tiny difference can be collected from the redundant part by cheap $1\times 1$ kernels. Then we fuse these two categories of extracted information accordingly to ensure no details get lost in a parameter-free way. Moreover, we design our SPConv in a generic way and make it a plug-and-play replacement for the vanilla convolution without any other adjustment to network architectures or hyper-parameter settings. Since our SPConv tolerates the redundancy among feature maps, it is orthogonal and complementary to current methods that focus on building better CNN topology \cite{huang2017densely,szegedy2015going,gao2019res2net}, reducing redundancy on channel \cite{xie2017aggregated,howard2017mobilenets} or spatial dimension \cite{DBLP:conf/iccv/Chen0XYKRYF19} and reducing redundancy in dense model parameters \cite{DBLP:conf/iclr/GaoZDMX19}. We can aggregate some of them reasonably to get a more lightweight model.

To the best of our knowledge, this is the first convolution/filter that 
adopts $k\times k$ kernels only on part of the input channels. All existing methods before, including HetConv \cite{singh2019hetconv} and OctConv \cite{DBLP:conf/iccv/Chen0XYKRYF19}, perform expensive $k\times k$ kernels on all input channels. Without any bells and whistles, experiments on current benchmarks demonstrate that in such a novel way, not only does our SPConv reduce the redundancy in vanilla convolutional operations, but also loses little useful information. Experimental SPConv-equipped architectures consistently outperform the state-of-the-art baselines in both accuracy and GPU inference time with nearly 60\%-70\% of the parameters and FLOPs dropped at most, which is superior to current works that are either (much) slower on GPU or less accurate, though they have pretty good parameters and FLOPs.

In conclusion, we make the following contributions:

- We reveal an overlooked redundancy in vanilla convolution and propose to split all input channels into two parts: one for the representative contributing intrinsic information and the other for the redundant complementing different details.

- We design a plug-and-play SPConv module to replace the vanilla convolution without any adjustment, which actually outperforms state-of-the-art baselines in both accuracy and inference time with parameters and FLOPs dropped sharply.


\section{Related Work}
\begin{figure*}
  \centering
  \includegraphics[width = 16.5cm]{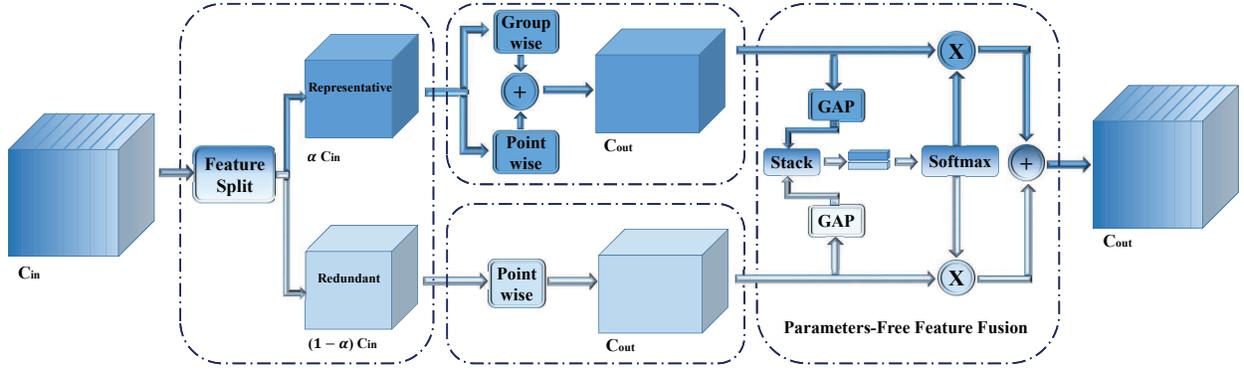}
  \caption{SPConv module.}\label{structure}
\end{figure*}
\subsection{Efficient and Compact Model Design}
There has been raising interest for researchers to create compact while accurate models since LeCun in 1990 \cite{lecun1990optimal}. Inception \cite{szegedy2015going} adopts split-transform-merge strategies and achieves low theoretical complexity with compelling accuracy. Meanwhile, some novel convolutional filters like Group-wise Convolution (GWC) \cite{krizhevsky2012imagenet}, Depth-wise Convolution (DWC) \cite{vanhoucke2014learning} and Point-wise Convolution (PWC) \cite{DBLP:journals/corr/LinCY13,DBLP:conf/cvpr/HeZRS16} are widely used for efficient model design. Evolved from Inception, ResNeXt \cite{xie2017aggregated} shares the same topology in all aggregated transformations with GWC and PWC, which reduces redundancy in inter-channel connectivity and introduces a new dimension called cardinality. To reduce connection density further, Xception \cite{DBLP:conf/cvpr/Chollet17} and MobileNet \cite{howard2017mobilenets} use DWC for spatial information extraction and PWC for channel information fusion successively. More thoroughly, ShuffleNet \cite{zhang2018shufflenet} adopts GWC on 1x1 convolutions followed by channel shuffle operation in addition to DWC on 3x3 convolutions.

Orthogonal to channel redundancy, OctConv \cite{DBLP:conf/iccv/Chen0XYKRYF19} explores redundancy on the spatial dimension of feature maps. Except for homogeneous convolution, HetConv \cite{singh2019hetconv} designs heterogeneous convolutional filters where exist both $3\times 3$ kernels and $1\times 1$ kernels in one single filter. Then these two heterogeneous kernels in filters of a particular layer are arranged in a shifted manner. With attention models being effective, SKNet \cite{Li_2019_CVPR} introduces a dynamic kernel selection mechanism based on multiple scales of input information and achieves adaptive receptive field.  The common part of these efficient models above, including OctConv \cite{DBLP:conf/iccv/Chen0XYKRYF19}, HetConv \cite{singh2019hetconv} and SKNet \cite{Li_2019_CVPR}, is that they apply $3\times 3$ kernels to all input channels. However, as shown in Figure \ref{panda}, many features maps appear much pattern similarity, which reveals vanilla convolutions are basically repetitive. Therefore, we propose SPConv module to alleviate this redundancy. With the idea similar to us, GhostNet \cite{ghostnet} also points out redundancy among feature maps. It uses ordinary convolution to generate some intrinsic features first, then utilizes cheap linear operation to do feature augmentation. Different from \cite{ghostnet}, our SPConv ensures the model grasp every original feature, instead of features generated by the intrinsic, so that we achieve better performance.

As for inference time on GPU, nearly all efficient models above are (much) slower than baseline.
Superior to them, our SPConv achieves faster inference speed than baseline.
\section{Approach}

In this section, we first introduce our proposed SPConv for cutting down the redundancy between feature maps,
then we analyze this module in detail.

\subsection{Vanilla Convolution}
Let $X\in R^{L \times h \times w}$, $Y\in R^{M\times h \times w}$ denote the input and convolved output tensors with $L$ input channels and $M$ output channels respectively. In general, square $k\times k$ convolutional kernels, denoted by $W \in R^{L\times k \times k \times M}$, are used to convolve $L$ input channels into $M$ output channels for feature extraction, resulting $Y=WX+b$. To simplify the notation, we omit the bias term and present the discussion over a single spatial position, say $k\times k$ area that a square kernel covers. The formulation for the complete $h\times w$ feature map can be obtained easily in the same way. So a vanilla convolution can be given as equation \ref{eq:1}:
{
\begin{equation} \label{eq:1}
\resizebox{.91\linewidth}{!}{$
\left[ \begin{array}{c}
{{y}_{1}} \\ {{y}_{2}} \\ {\vdots} \\ {{y}_{{M}}}\end{array}\right]
=
\left[\hspace{-0.15cm}\begin{array}{cccc}
{{W}_{11}} &  {{W_{12}}} & {\cdots} & {{W_{1,{L}}}}\\
  {{W_{21}}}&{{W}_{22}} & {{\cdots}} & {{W_{2,{L}}}} \\
  {\vdots} & {\vdots} & {\ddots} & {\vdots} \\
  {{W_{{M},1}}} & {{{W}_{{{M}},{2}}}} & {{\cdots}} & {{W}_{{M},{L}}}\end{array}\right]
  \left[\begin{array}{c}{{x}_{1}} \\ {{x}_{2}} \\ {\vdots} \\ {{x}_{{L}}}\end{array}\right]
$}
\end{equation}
}
where $x_{i}, i\in [1,L]$ is one of the $L$ input  matrixes, and $W_{ij}, i,j \in [1; L, M]$ is one of the parameters of $M$ filters, both of which are ($k\times k$) - dimensional. After convolution, we get $M$ convolved outputs, denoted by $y_{j}, j \in [1,M]$.\\
\subsection{The Representative and the Redundant}
All existing filters, such as vanilla convolution, GhostConv \cite{ghostnet}, OctConv \cite{DBLP:conf/iccv/Chen0XYKRYF19} and HetConv \cite{singh2019hetconv}, perform $k\times k$ convolution on all input channels. However, as shown in Figure \ref{panda}, there is intuitive pattern similarity among intermediate feature maps so that some $k\times k$ convolutions are relatively redundant. Meanwhile, there are also no two identical channels so that we can not throw away these redundant channels neither. Inspired by this idea, our SPConv splits all input channels into two main parts in a ratio of $\alpha$, one for the representative applying $k\times k$ (here we adopt widely used $3\times 3$) convolution to supply intrinsic information; the other for the redundant applying cheap $1\times 1$ convolution to complement tiny hidden details as shown in the left part of Figure \ref{structure}. So our initial SPConv can be given as equation \ref{eq:2}:
{\begin{equation}\label{eq:2}
\resizebox{.91\linewidth}{!}{$
\hspace{-0.2cm}\left[\hspace{-0.1cm}\begin{array}{c}
{{y}_{1}} \\ {{y}_{2}} \\ {\vdots} \\ {{y}_{{M}}}\end{array}\hspace{-0.15cm}\right] \hspace{-0.15cm}=
{\hspace{-0.15cm}\left[\hspace{-0.22cm}\begin{array}{ccc}
{{W_{11}}}\hspace{-0.4cm} & {{\cdots}}\hspace{-0.4cm} & {{W_{1,\alpha {L}}}}\\
{{\vdots}} \hspace{-0.4cm}& {{\ddots}}\hspace{-0.4cm} & {{\vdots}} \\
{{W_{{M},1}}}\hspace{-0.4cm} & {{\cdots}}\hspace{-0.4cm} & {{W_{{M},\alpha {L}}}}\\
\end{array}\hspace{-0.22cm}\right]}
{\hspace{-0.12cm}\left[\hspace{-0.2cm}\begin{array}{c}
{{x_1}} \\ {{\vdots}} \\ {{x_{\alpha {L}}}}
\end{array}\hspace{-0.18cm}\right]}
\hspace{-0.1cm}+\hspace{-0.1cm}
{\left[\hspace{-0.2cm}\begin{array}{ccc}
{{w_{1,\alpha {L}+1}}} \hspace{-0.3cm}& {{\cdots}}\hspace{-0.2cm}  &\hspace{-0.1cm} {{w_{1,{L}}}}\\
{{\vdots}}\hspace{-0.3cm}  & {{\ddots}}\hspace{-0.2cm} & \hspace{-0.1cm} {{\vdots}} \\
{{w_{{M},\alpha {L}+1}}} \hspace{-0.3cm}& {{\cdots}} \hspace{-0.2cm} & \hspace{-0.1cm}{{w_{{M},{L}}}}\\
\end{array}\hspace{-0.2cm} \right]}
{\hspace{-0.1cm} \left[\hspace{-0.2cm} \begin{array}{c}
{{x_{\alpha {L}+1}}} \\ {{\vdots}} \\ {{x_{{L}}}}
\end{array}\hspace{-0.18cm} \right]\hspace{-0.2cm} }
$}
\end{equation}}
where ${W}_{i,j}, j\in [1, \alpha L]$ represents parameters of $3\times 3$ kernels convolving on the $\alpha L$ representative channels. ${w}_{i,j}, j\in [\alpha L +1, L]$ stands for parameters of cheap $1\times 1$ kernels performing on the remaining $(1-\alpha)L$ redundant features with pointwise convolution. HetConv \cite{singh2019hetconv} also adopts this heterogeneous operation similar to us, while we share every filter with the same heterogeneous distribution instead of a shifted manner,whose fragmented kernels make computation more difficult on current hardware, so that we can achieve faster inference speed.
\subsection{Further Reduction for the Representative}
After splitting all input channels into two main parts, there might be redundancy among the representative part. In other words, representative channels can be divided into several parts and each part stands for one main category of features, e.g., color and textures. Thus we adopt group convolution on the representative channels to reduce redundancy further as shown in the middle part of Figure \ref{structure}. We can view a group convolution as a vanilla convolution with a sparse block-diagonal convolutional kernel, where each block corresponds to a partition of channels and there are no connections across the partitions \cite{zhang2017interleaved}. That means, after group convolution, we reduce redundancy among the representative part further while we also cut off the connection across channels which might be useful inevitably. We remedy this information loss by adding pointwise convolution across all representative channels. Different from the commonly used group convolution followed by pointwise convolution, we conduct both GWC and PWC on the same representative channels. Then we fuse the two resulted features by direct summation because of their same channel origin, which obtains an extra score (here we set group size as 2). So the representative part of the equation \ref{eq:2} can be formulated as equation \ref{eq:3}:
{\begin{equation}\label{eq:3}
\resizebox{.91\linewidth}{!}{$
\left[\hspace{-0.1cm}\begin{array}{ccc}
{{W}_{11}^{p}} & {0} & {0}  \\
{0} & {\ddots} & {0} \\
{0} & {0} & {{W}_{G G}^{p}}
\end{array}\right]\hspace{-0.1cm}
\left[\begin{array}{c}{{z}_{1}} \\ {\vdots} \\ {{z}_{G}}\end{array}
\right]+
{\left[\begin{array}{ccc}
{{w_{11}}} & {{\cdots}} & {{w_{1,\alpha {L}}}}\\
{{\vdots}} & {{\ddots}} & {{\vdots}} \\
{{w_{{M},1}}} & {{\cdots}} & {{w_{{M},\alpha {L}}}}\\
\end{array}\right]}
{\hspace{-0.1cm}\left[\begin{array}{c}
{{x_1}} \\ {{\vdots}} \\ {{x_{\alpha {L}}}}
\end{array}\right]}
$}
\end{equation}
}

Here we divide the ${\alpha c_L}$  representative channels into G groups and each group ${z}_l$ contains (${\alpha c_L/G}$) channels. ${W}^p_{ll}$ are parameters of group convolutional kernel in the $l$th group.
\subsection{Parameter-Free Feature Fusion Module}
So far, we have splitted the vanilla $3\times 3$ convolution into two operations: for the representative part, we conduct direct summation fusion of $3\times 3$ group convolution and $1\times 1$ pointwise convolutions to countervail grouped information loss; for the redundant part, we apply $1\times 1$ kernels to complement some tiny useful details. As a result, we get two categories of features. Because these two features originate from different input channels, a fusion method is needed to control information flow. Different from direct summation fusion as equation \ref{eq:2}, we design a novel feature fusion module for our SPConv, without extra parameters imported and being helpful to achieve better performance. As shown in the right part of Figure \ref{structure}, we use global average pooling to generate channel-wise statistics $S_1$, $S_3$ $\in R^C$ for global information embedding. By shrinking output features $U$ through its spatial dimension, the c-th element of $S$ can be given as equation \ref{eq:45}:
{
\begin{align}\label{eq:45}
{S_{kc}} = {F_{gap}(U_{kc})=\frac{1}{H \times W} \sum_{i=1}^{H} \sum_{j=1}^{W} {U}_{kc}(i, j)}, k\in {[1,3]}
\end{align}
}
Then we stack these two resulted $S_3$, $S_1$ vectors together followed by soft attention operation across channels to generate feature importance vector ${\beta}\in R^c$, ${\gamma} \in R^c$, the c-th element of which are given as follows:
{\begin{equation}\label{eq:eq6}
{\beta_{c}=\frac{e^{S_{3c}}}{e^{S_{3c}}+e^{S_{1c}}}},{\gamma_c = 1-\beta_c}
\end{equation}
}
Our final output ${Y}$ can be obtained by fusing features ${U_3}$ and ${U_1}$, which are from the representative and the redundant part respectively, directed by the feature importance vector $\beta$, $\gamma$ in a channel-wise manner.
{\begin{equation}\label{eq:7}
{Y}={\beta U_3} + {\gamma U_1}
\end{equation}
}
In summary, the vanilla convolution ${Y=WX}$ (bias term omitted) can be approximated by ${Y=W'X}$ as equation \ref{eq:8}:
{
\begin{align}\label{eq:8}
\nonumber
{Y} =&{W'X} \\ \nonumber
\approx &
\beta\left[\hspace{-0.15cm}\begin{array}{ccc}
{{W}_{11}^{p}}\hspace{-0.1cm} & \hspace{-0.1cm}{0}\hspace{-0.2cm} & \hspace{-0.1cm}{0}\hspace{-0.1cm}  \\
{0}\hspace{-0.2cm} & \hspace{-0.1cm}{\ddots} \hspace{-0.1cm}& \hspace{-0.1cm}{0} \hspace{-0.1cm}\\
{0}\hspace{-0.2cm} & \hspace{-0.1cm}{0}\hspace{-0.1cm} & \hspace{-0.1cm}{{W}_{G G}^{p}}\hspace{-0.1cm}
\end{array}\hspace{-0.1cm}\right]\hspace{-0.15cm}
\left[\hspace{-0.2cm}\begin{array}{c}{{z}_{1}} \\ {\vdots} \\ {{z}_{G}}\end{array}\hspace{-0.2cm}
\right]+
\beta{\left[\hspace{-0.2cm}\begin{array}{ccc}
{{w_{11}}} & \hspace{-0.2cm}{{\cdots}} & \hspace{-0.2cm}{{w_{1,\alpha {L}}}}\\
{{\vdots}} & \hspace{-0.2cm}{{\ddots}} & \hspace{-0.2cm}{{\vdots}} \\
{{w_{{M},1}}} & \hspace{-0.2cm}{{\cdots}} &\hspace{-0.2cm} {{w_{{M},\alpha {L}}}}\\
\end{array}\hspace{-0.2cm}\right]}
{\hspace{-0.1cm}\left[\hspace{-0.2cm}\begin{array}{c}
{{x_1}} \\ {{\vdots}} \\ {{x_{\alpha {L}}}}
\end{array}\hspace{-0.2cm}\right]} \\
&+\hspace{-0.05cm}{{\gamma}}
{\left[\hspace{-0.2cm}\begin{array}{ccc}
{{w_{1,\alpha L+1}}}\hspace{-0.25cm} & {{\cdots}}\hspace{-0.25cm} & {{w_{1,L}}}\\
{{\vdots}} \hspace{-0.25cm}& {{\ddots}} \hspace{-0.25cm}& {{\vdots}} \\
{{w_{M,\alpha L+1}}}\hspace{-0.25cm} & {{\cdots}} \hspace{-0.25cm}& {{w_{M,L}}}\\
\end{array}\hspace{-0.15cm}\right]}
{\hspace{-0.1cm}\left[\hspace{-0.2cm}\begin{array}{c}
{{x_{\alpha L+1}}} \\ {{\vdots}} \\ {{x_{L}}}
\end{array}\hspace{-0.2cm}\right]}
\end{align}}
where uppercase ${W}$ represents $3\times 3$ kernels and lowercase ${w}$ represents $1\times 1$ kernels. By replacing vanilla weights $W$ with the proposed SPConv weights $W'$, classic network architectures can obtain better performance in both accuracy and inference time on GPU with FLOPs and parameters dropped sharply as shown in section 4. Module decomposition tests of our SPConv are conducted in section 4.4.
\subsection{Complexity Analysis}
From equation \ref{eq:1}, the parameters of vanilla convolution can be calculated as:
{\begin{equation}
P_{va} = k \times k \times L \times M
\end{equation}}
From equation \ref{eq:8}, our SPConv has parameters of:
{\begin{align}
\nonumber
P_{sp} =& k \times k \times \frac{1}{g} \alpha {L} \times \frac{1}{g}{M} \times g + 1\times 1 \times \alpha L \times {M}\\ \nonumber
&+ 1\times 1 \times(1-\alpha) {L} \times {M} \\
=&(k \times k \times \frac{\alpha}{g}+1)\times {L} \times {M}
\end{align}}
When we choose half of the input channels as the representative, divide the representative part into 2 groups and adopt the widely used 3x3 kernels, the amount of parameters can be reduced by 2.8 times while the model actually achieves a little better performance than vanilla convolution in both accuracy and inference speed.
\section{Experiment}
To show the effectiveness of the proposed SPConv, in this section, we conduct experiments with only the widely-used $3\times 3$ kernels being replaced by our SPConv modules. The upgraded networks have only one global hyper-parameter $\alpha$, denoting the ratio of the representative features among all input features. Although different layers/stages might require different number of representative features, we use the same $\alpha$ for all layers in the experiments. Carefully designing different $\alpha$ for different layers would find a more compact model. Meanwhile, we set group size for the representative part as 2.

Firstly, we perform small scale image classification experiments on the CIFAR-10 dataset \cite{krizhevsky2009learning} with ResNet-20 \cite{DBLP:conf/cvpr/HeZRS16} and VGG-16 \cite{DBLP:journals/corr/SimonyanZ14a} architectures. Then we experiment a large scale 1000-class single label classification task on ImageNet-2012 \cite{DBLP:conf/cvpr/DengDSLL009} with ResNet-50 \cite{DBLP:conf/cvpr/HeZRS16} architecture. To explore SPConv's generality further, we also conduct a multi-label object detection experiment on MS COCO dataset \cite{lin2014microsoft}. For fair comparisons, all models in each experiment, including re-implemented baselines and SPConv-equipped models, are trained from scratch on 4 NVIDIA Tesla V100 GPUs with the default data augmentation and training strategy which are optimized for vanilla convolution and no other tricks are used. Therefore, our proposed SPConv may achieve better performance with extensive hyper-parameter searches. More ablation studies are performed on small scale CIFAR-10 dataset.

\subsection{Small Scale Classfication Task}

\subsubsection{VGG-16 on CIFAR-10}

As we know, VGG-16 \cite{DBLP:journals/corr/SimonyanZ14a} is originally designed for 1000-label ImageNet with 13 conv layers and 3 fully-connected layers. For 10-label CIFAR, we choose its widely used variant: VGG-15 with 2 fully-connected layers and batch normalization \cite{DBLP:conf/icml/IoffeS15} equipped after each layer. Same as \cite{singh2019hetconv}, we only replace the last 12 $3\times 3$ conv layers with our proposed SPConvs except the initial $3\times 3$ conv layer and keep all other configurations unchanged. Optimization is performed using SGD with weight decay = 5e-4, batch-size = 128, initial learning rate = 0.1 which is divided by 10 every 50 epochs.

As shown in Table \ref{CIFAR}, when we set the split ratio $\alpha$ from 1 to 1/16 gradually, the FLOPs and parameters decrease sharply without any significant accuracy decrease. Without any bells and whistles, our proposed SPConv does a better job with only 33\% of FLOPs and parameters when the split ratio is set as 1/2. That is, although fewer input channels are used for intrinsic information extraction, the model is still robust enough. That also means some input channels, or feature maps, of a conv layer are redundant and have no need to be counted into computation by expensive $3\times 3$ kernels. Compared to related works \cite{singh2019hetconv,ghostnet} on VGG-16, our SPConv outperforms Ghost-Conv and HetConv. We argue that, for a specific task or dataset, representative and intrinsic features count the most, while other redundant or similar features can be saved using cheap $1\times 1$ kernels to complement tiny different details. To challenge more, we perform experiments on the much more compact ResNet-20 \cite{DBLP:conf/cvpr/HeZRS16}.
\begin{table}[]\center
\tiny{
\begin{tabular}{|c|c|c|c|c|c|}
\hline
\multicolumn{6}{|c|}{CIFAR\_10 - ResNet\_20}                                                                                   \\ \hline
Model                         & FLOPs            & \makecell[c]{FLOPs\\Reduced}    & Params     & \makecell[c]{Params\\Reduced} & ACC@1            \\ \hline
ResNet\_20-Baseline                 & 41.62M           & -                & 0.27M          & -                  & 92\%             \\ \hline
\textbf{SPConv-ResNet\_20-$\alpha$1/2}        & \textbf{17.91M}           & \textbf{56.96\%}          & \textbf{0.10M}          & \textbf{63.00\%}            & \textbf{92.23\%}          \\ \hline
\textbf{SPConv-ResNet\_20-$\alpha$1/4}        & \textbf{12.89M}          & \textbf{75.88\%}          & \textbf{0.071M}        & \textbf{73.70\%}            & \textbf{91.15\%}          \\ \hline
\multicolumn{6}{|c|}{CIFAR\_10 - VGG\_16}                                                                                      \\ \hline
VGG\_16-Baseline              & 349.51M          & -                & 16.62M         & -                  & 94.00\%          \\ \hline
Ghost-VGG\_16-s2              & 158M             & 45.20\%          & 7.7M           & 53.67\%            & 93.70\%          \\ \hline
\textbf{SPConv-VGG\_16-$\alpha$1/2}  & \textbf{118.27M} & \textbf{66.24\%} & \textbf{5.6M}  & \textbf{66.30\%}   & \textbf{94.40\%} \\ \hline
HetConv-VGG\_16-P4            & 105.98M          & 69.67\%          & 5.17M          & 68.89\%            & 93.92\%          \\ \hline
\textbf{SPConv-VGG\_16-$\alpha$1/4}  & \textbf{79.34M}  & \textbf{77.29\%} & \textbf{3.77M} & \textbf{77.31\%}   & \textbf{93.94\%} \\ \hline
HetConv-VGG\_16-P8            & 76.89M           & 78.00\%          & 3.54M          & 78.70\%            & 93.86\%          \\ \hline
\textbf{SPConv-VGG\_16-$\alpha$1/8}  & \textbf{59.87M}  & \textbf{82.87\%} & \textbf{2.85M} & \textbf{82.85\%}   & \textbf{93.77\%} \\ \hline
\textbf{SPConv-VGG\_16-$\alpha$1/16} & \textbf{55.14M}  & \textbf{84.22\%} & \textbf{2.39M} & \textbf{85.62\%}   & \textbf{93.43\%} \\ \hline
\end{tabular}
}
\tiny{\caption{Results for ResNet-20 and VGG-16 on CIFAR-10.}\label{CIFAR}}
\end{table}
\subsubsection{ResNet-20 on CIFAR-10}
ResNet-20 is composed of three stages of convolutional layers with each stage containing 16-32-64 filters respectively, whose features are much fewer than 64-128-256-512's VGG-16. ResNet-20 contains only 0.27M parameters, about 1.6\% of VGG-16, while it performs only 2\% less accurate. It would be much more challenging for our SPConv to slim such a compact architecture. As settings the same as VGG-16 above, we replace all the $3\times 3$ vanilla conv layers with our proposed SPConvs except the initial one, adopt default training strategy as \cite{DBLP:conf/cvpr/HeZRS16} and re-implement faithfully. The results are shown in Table \ref{CIFAR}.

From Table \ref{CIFAR}, we can see the proposed SPConv slims such a compact ResNet-20 successfully and the performance is almost the same as with VGG-16 before. When we take the first half of channels (8-16-32 in each stage) for intrinsic information extraction and complement tiny different details with the remaining half of channels using $1\times 1$ kernels, the FLOPs and the parameters drops about 60\% while the slimmed model still performs a little better than the original one. To go slimmer, we set the split ratio as 1/4, which means, except the initial conv layer, only 4 channels in the first stage, 8 channels in the second stage and 16 channels in the third stage are used to extract intrinsic information. The results show that our slimmer model performs only 0.75\% worse than the baseline, which is as expected. Because under such a severe condition, $3\times 3$ kernels have not enough power to count more intrinsic features in and $1\times 1$ kernels have poor ability to extract more information so that there are not enough representative and intrinsic features which are important for the model to classify the dataset correctly. To be more convincing, we also conduct large scale experiments on ImageNet with ResNet-50.

\subsection{Large Scale Experiments on ImageNet}
\begin{table*}[]\center
\scriptsize
\begin{tabular}{|c|c|c|c|c|c|c|c|}
\hline
\multicolumn{8}{|c|}{ImageNet2012-ResNet50}                                                                                                                      \\ \hline
Model                         & FLOPs         & \makecell[c]{FLOPs\\Reduced}    & Params          & \makecell[c]{Params\\Reduced}   & Acc@1            & Acc@5            & \makecell[c]{Inference Time\\on GPU}    \\ \hline
ResNet50-Baseline             & 4.14G          & -                & 25.56M          & -                & 75.91\%          & 92.78\%          & 1.32 ms          \\ \hline
\textbf{SPConv-ResNet50-$\alpha$1/2} & \textbf{2.97G} & \textbf{28.26\%} & \textbf{18.34M} & \textbf{28.24\%} & \textbf{76.26\%} & \textbf{93.05\%} & \textbf{1.23 ms} \\ \hline
HetConv-ResNet50-P4           & 2.85G          & 30.32\%          & -               & -                & 76.16\%          & -                & -                 \\ \hline
\textbf{SPConv-ResNet50-$\alpha$1/4} & \textbf{2.74G} & \textbf{33.82\%} & \textbf{16.93M} & \textbf{33.76\%} & \textbf{75.95\%} & \textbf{92.99\%} & \textbf{1.19 ms} \\ \hline
\textbf{SPConv-ResNet50-$\alpha$1/8} & \textbf{2.62G} & \textbf{36.72\%} & \textbf{16.22M} & \textbf{36.54\%} & \textbf{75.40\%} & \textbf{92.77\%} & \textbf{1.17 ms} \\ \hline
OctConv-ResNet50-$\alpha$0.5$^\dagger$         & 2.40G          & 42.00\%          & 25.56M          & 0.00\%           & 76.40\%          & 93.14\%          & 3.51 ms          \\ \hline
Ghost-ResNet50-s2            & 2.20G          & 46.85\%          & 13.0M           & 49\%             & 75\%             & 92.3\%           & -                 \\ \hline
\end{tabular}
\caption{shows large scale experiments for ResNet-50 on ImageNet-2012 in different split ratios with group-size = 2.}\label{ImageNet}
\end{table*}

\begin{table*}[] \center
\scriptsize
\setlength{\tabcolsep}{2.82mm}{
\begin{tabular}{|ccccccccc|}
\hline
\multicolumn{9}{|c|}{Object Detection Experiments on MS COCO}                                                                                                                                                                                                                                                                                                                                                                                                                                                                                                                                                                             \\ \hline
\multicolumn{1}{|c|}{Framework}                     & \multicolumn{1}{c|}{\begin{tabular}[c]{@{}c@{}}Backbone\\     (ResNet-50)\end{tabular}} & \multicolumn{1}{c|}{\begin{tabular}[c]{@{}c@{}}Backbone\\     Params/FLOPs\end{tabular}} & \multicolumn{1}{c|}{-}                       & \multicolumn{1}{c|}{\begin{tabular}[c]{@{}c@{}}AP@\\     IoU=0.5\end{tabular}} & \multicolumn{1}{c|}{\begin{tabular}[c]{@{}c@{}}Small\\     objects\end{tabular}} & \multicolumn{1}{c|}{\begin{tabular}[c]{@{}c@{}}Medium\\     objects\end{tabular}} & \multicolumn{1}{c|}{\begin{tabular}[c]{@{}c@{}}Large\\     objects\end{tabular}} & All  \\ \hline
\multicolumn{1}{|c|}{\multirow{4}{*}{Faster-R-CNN}} & \multicolumn{1}{c|}{Vanilla}                                                            & \multicolumn{1}{c|}{23.51M/4.13G}                                                        & \multicolumn{1}{c|}{\multirow{2}{*}{AP(\%)}} & \multicolumn{1}{c|}{58.2}                                                      & \multicolumn{1}{c|}{21.6}                                                        & \multicolumn{1}{c|}{39.8}                                                         & \multicolumn{1}{c|}{46.8}                                                        & 36.3 \\ \cline{2-3} \cline{5-9}
\multicolumn{1}{|c|}{}                              & \multicolumn{1}{c|}{\textbf{SPConv}}                                                             & \multicolumn{1}{c|}{\textbf{16.29M/2.96G}}                                                        & \multicolumn{1}{c|}{}                        & \multicolumn{1}{c|}{\textbf{58.9}}                                                      & \multicolumn{1}{c|}{\textbf{21.8}}                                                        & \multicolumn{1}{c|}{\textbf{40.4}}                                                         & \multicolumn{1}{c|}{\textbf{46.3}}                                                        & \textbf{36.4} \\ \cline{2-9}
\multicolumn{1}{|c|}{}                              & \multicolumn{1}{c|}{Vanilla}                                                            & \multicolumn{1}{c|}{23.51M/4.13G}                                                        & \multicolumn{1}{c|}{\multirow{2}{*}{AR(\%)}} & \multicolumn{1}{c|}{-}                                                         & \multicolumn{1}{c|}{32.7}                                                        & \multicolumn{1}{c|}{55.3}                                                         & \multicolumn{1}{c|}{64.5}                                                        & 51.5 \\ \cline{2-3} \cline{5-9}
\multicolumn{1}{|c|}{}                              & \multicolumn{1}{c|}{\textbf{SPConv}}                                                             & \multicolumn{1}{c|}{\textbf{16.29M/2.96G}}                                                        & \multicolumn{1}{c|}{}                        & \multicolumn{1}{c|}{-}                                                         & \multicolumn{1}{c|}{\textbf{33.9}}                                                        & \multicolumn{1}{c|}{\textbf{55.6}}                                                         & \multicolumn{1}{c|}{\textbf{63.7}}                                                        & \textbf{51.6} \\ \hline
                                                    &                                                                                         &                                                                                          &                                              &                                                                                &                                                                                  &                                                                                   &                                                                                  &      \\ \hline
\multicolumn{1}{|c|}{\multirow{4}{*}{RetinaNet}}    & \multicolumn{1}{c|}{Vanilla}                                                            & \multicolumn{1}{c|}{23.51M/4.13G}                                                        & \multicolumn{1}{c|}{\multirow{2}{*}{AP(\%)}} & \multicolumn{1}{c|}{55.6}                                                      & \multicolumn{1}{c|}{19.7}                                                        & \multicolumn{1}{c|}{39.3}                                                         & \multicolumn{1}{c|}{47}                                                          & 35.6 \\ \cline{2-3} \cline{5-9}
\multicolumn{1}{|c|}{}                              & \multicolumn{1}{c|}{\textbf{SPConv}}                                                             & \multicolumn{1}{c|}{\textbf{16.29M/2.96G}}                                                        & \multicolumn{1}{c|}{}                        & \multicolumn{1}{c|}{\textbf{56.4}}                                                      & \multicolumn{1}{c|}{\textbf{20.6}}                                                        & \multicolumn{1}{c|}{\textbf{40.4}}                                                         & \multicolumn{1}{c|}{\textbf{47.5}}                                                        & \textbf{35.9} \\ \cline{2-9}
\multicolumn{1}{|c|}{}                              & \multicolumn{1}{c|}{Vanilla}                                                            & \multicolumn{1}{c|}{23.51M/4.13G}                                                        & \multicolumn{1}{c|}{\multirow{2}{*}{AR(\%)}} & \multicolumn{1}{c|}{-}                                                         & \multicolumn{1}{c|}{32.9}                                                        & \multicolumn{1}{c|}{56.7}                                                         & \multicolumn{1}{c|}{66.9}                                                        & 52.2 \\ \cline{2-3} \cline{5-9}
\multicolumn{1}{|c|}{}                              & \multicolumn{1}{c|}{\textbf{SPConv}}                                                             & \multicolumn{1}{c|}{\textbf{16.29M/2.96G}}                                                        & \multicolumn{1}{c|}{}                        & \multicolumn{1}{c|}{-}                                                         & \multicolumn{1}{c|}{\textbf{33.4}}                                                        & \multicolumn{1}{c|}{\textbf{56.8}}                                                         & \multicolumn{1}{c|}{\textbf{68.1}}                                                        & \textbf{52.8} \\ \hline
\end{tabular}}
\caption{shows the multi-label object detection results for Faster-R-CNN and RetinaNet on MS COCO with $\alpha=\frac{1}{2}.$}\label{MSCOCO}
\end{table*}
We conduct large scale experiments with ResNet-50 \cite{DBLP:conf/cvpr/HeZRS16} on the most authoritative ImageNet. As before, we replace standard $3\times 3$ kernels with the proposed SPConv except for the first 7x7 conv layer and keep all other parts unchanged. Due to limited GPU resource and training time expense, we choose NVIDIA DALI\footnote{\url{https://github.com/NVIDIA/DALI}} project as our baseline to reimplement ResNet-50 faithfully. DALI differs in pre-processing image data on GPU for data loading acceleration and using apex \cite{DBLP:conf/iclr/MicikeviciusNAD18} for mixed-precision fp16/fp32 to speed up training without significant accuracy drop. With the default settings, the learning rate starts at 0.1 and decays by a factor of 10 every 30 epochs, using synchronous SGD with weight decay 1e-4, momentum 0.9 and a mini-batch of 256 to train the model from scratch for 90 epochs.

From Table \ref{ImageNet}, we can see our SPConv still performs 0.27\% better than baseline when we set the split ratio as 1/2 and performs almost the same when $\alpha$ is set as 1/4 with FLOPs reduced about 28\% and 34\% separately in such a huge dataset. The reason why drop ratios of params and FLOPs there is not as high as VGG-16 or ResNet-20 above is attributed to unreplaced initial 7x7 Conv and more fully-connected-layer parameters. When we decrease $\alpha$ to 1/8 continuously, FLOPs and parameters drop by about 37\% and the Top1-Acc is about 0.6\% less accurate, which means 8-16-32-64 representative and intrinsic features in each stage have been in shortage for ResNet-50 to classify such a huge ImageNet correctly.

Inference time is tested on a single NVIDIA Tesla V100 with NVIDIA DALI as data pipelines. Batch size is set as 64. For current efficient convolution designs, though they have fewer parameters and FLOPs, most of them are at the expense of (much) lower inference speed on GPU or a little worse accuracy than baselines. Superior to them, our SPConv outperforms the baseline in both inference time and accuracy. For fair comparisons, we reimplement OctConv with standard training strategy as \protect\cite{DBLP:conf/cvpr/HeZRS16} and no other tricks in \protect\cite{DBLP:conf/iccv/Chen0XYKRYF19} are used. Although the results show that Octave performs a little better than our SPConv in accuracy with fewer FLOPs, Octave requires as many parameters as the baseline. For inference time, our SPConv outperforms OctConv sharply on GPU.

\subsection{Object Detection on MS COCO}

To evaluate our SPConv's generalizability further, we test our SPConv on the object detection task of MS COCO \cite{lin2014microsoft}, using two-stage Faster R-CNN[15] and one-stage RetinaNet \cite{lin2017focal} framework respectively with ResNet-50-FPN \cite{DBLP:conf/cvpr/HeZRS16,lin2017feature} backbone as baseline. For fair comparisons, experiments are implemented on mmdection \cite{DBLP:journals/corr/abs-1906-07155} where we keep all other settings default except replacing vanilla $3\times 3$ conv layers with our proposed SPConv ($\alpha$=$\frac{1}{2}$, g=2) in ResNet-50 backbone and use pre-trained backbone parameters in section 4.3. Similar to \cite{DBLP:conf/iccv/HeGDG17}, models are trained on the COCO trainval35k set and tested on the left 5K minival set.

Table \ref{MSCOCO} shows the comparisons in mean average precision (mAP) and mean average recall (mAR). Our SPconv-equipped Faster R-CNN and RetinaNet outperforms their baseline respectively. SPConv-equipped Faster R-CNN performs a little worse for large object, and we attribute this to our replacement of some expensive $3\times 3$ kernels whose receptive field is large with cheap $1\times 1$ kernels whose receptive field is small.
\subsection{Ablation Studies}
From section 3, our SPConv consists of three main parts, 1) cheap $1\times 1$ convolution for the redundant channels, 2) $3\times 3$ GWC + PWC for the representative features and 3) parameter-free feature fusion method. In this section, we conduct an ablation study to investigate the relative effectiveness of each part in SPConv. All ablation experiments are conducted with ResNet-20 on CIFAR-10.

\subsubsection{Effectiveness of GWC+PWC}
We adopt GWC+PWC on the representative features. To evaluate its effectiveness, we experiment with two commonly used methods . The first one is GWC followed by PWC. As show in the second line of Table \ref{ablation}, our GWC+PWC has about 0.4\% better accuracy  with fewer FLOPs. The second one is vanilla $3\times 3$ convolution, which achieves the best accuracy, 0.04\% better while demanding 43\% more FLOPs.
\subsubsection{Parameters-Free Feature Fusion Method}
Compared with the proposed feature fusion method, we experiment with direct summation fusion. As show in the fourth line of Table \ref{ablation}, our parameters-free fusion method achieves 0.7\% better with only 0.35M more FLOPs.
\subsubsection{Count the Redundant Channels or Not}
We examine our naive thought: delete the redundant channels directly, without feature fusion accordingly. As shown in the last line of Table \ref{ablation}, performance drops about 3\%(though fewer parameters), which means redundant features contain useful tiny details and can not be deleted directly.
\begin{table}[t]{\scriptsize
\setlength{\tabcolsep}{1.65mm}{
\begin{tabular}{|c|c|c|c|c|c|c|c|c|}
\hline
                           & P1 & P20 & P21 & P22 & P3 & Acc(\%)   & FLOPs   & Params \\ \hline
\multirow{5}{*}{ResNet-20} & \textbf{\checkmark}  &     & \textbf{\checkmark}   &     & \textbf{\checkmark}  & \textbf{92.23}$\pm$ \textbf{0.1} & \textbf{17.91M}  & \textbf{0.10M}  \\ \cline{2-9}
                           & \checkmark  &     &     & \checkmark   & \checkmark  & 91.84$\pm$ 0.2 & 20.38M  & 0.12M  \\ \cline{2-9}
                           & \checkmark  & \checkmark   &     &     & \checkmark  & 92.27$\pm$ 0.1 & 25.70M  & 0.15M  \\ \cline{2-9}
                           & \checkmark  &     &  \checkmark   &    &    & 91.51$\pm$ 0.2 & 17.56M & 0.10M  \\ \cline{2-9}
                           &    &     & \checkmark   &     &    & 89.43$\pm$ 0.1 & 14.47M  & 0.088M \\ \hline
\end{tabular}}}
\footnotesize{\scriptsize
P1: Count the redundant channels or not;\\
P20: Use vanilla convolution for the representative; \\
P21: Use GWC+PWC for the representative;\\
P22: Use GWC followed by PWC for the representative;\\
P3: Parameter-free feature fusion method instead of direct summation fusion.}
\caption{ablation studies on CIFAR-10 in different setups.}\label{ablation}
\end{table}
\vspace{-0.08cm}
\section{Conclusion}
In this work, we revealed an overlooked pattern redundancy in vanilla convolution. To alleviate this problem, we proposed a novel SPConv module to split all input feature maps into the representative part and the uncertain redundant part. Extensive experiments on various datasets and network architectures demonstrated our SPConv's effectiveness. More importantly, it achieves faster inference speed than baseline, which is superior to existing works. Multi-label object detection experiments conducted on MS COCO proved its generality. Since our SPConv is orthogonal and complementary to the current model compression method, careful combinations of them would get a more lightweight model in the future.
\bibliographystyle{named}
\bibliography{ijcai20}
\end{document}